\documentclass[onefignum,onetabnum,a4,reqno]{siamart171218}

\usepackage[top=3cm, left=4cm, right=4cm, bottom=3cm]{geometry}

\usepackage{lipsum}
\usepackage{amsfonts}
\usepackage{graphicx}
\usepackage{epstopdf}
\usepackage{algorithmic}
\ifpdf
  \DeclareGraphicsExtensions{.eps,.pdf,.png,.jpg}
\else
  \DeclareGraphicsExtensions{.eps}
\fi

\usepackage{color}

\usepackage{amssymb}
\usepackage{amsmath}
\usepackage{bm}

\usepackage{scalerel} 

\usepackage{dutchcal}

\usepackage[shortcuts]{extdash}

\usepackage{mathrsfs} 

\usepackage{physics}

\usepackage{graphicx} 
\usepackage{epsfig}
\usepackage{ifpdf}
\usepackage{caption}

\usepackage{listings}

\newcommand{\R}[1]{\ensuremath{\mathbb{R}^{#1}}}

\newcommand{\vr}[1]{\ensuremath{\mathbf{#1}}}

\newcommand{\mt}[1]{\ensuremath{\mathbf{#1}}}

\newcommand{\q}{\ensuremath{\mathbf{q}}}

\newcommand{\dq}{\ensuremath{\mathbf{\dot{q}}}}

\newcommand{\ddq}{\ensuremath{\mathbf{\ddot{q}}}}

\newcommand{\G}{\ensuremath{\mathbf{G}}}

\newsiamremark{remark}{Remark}
\newsiamremark{hypothesis}{Hypothesis}
\crefname{hypothesis}{Hypothesis}{Hypotheses}
\newsiamthm{claim}{Claim}

\headers{Computed torque control: an introduction}{L. Ros}

\title{\small How to design, and tune, a computed torque controller: \\ {\normalfont  \footnotesize An introduction, and a Matlab example}}

\author{Llu\'{\i}s Ros\thanks{Institut de Robòtica i Informàtica Industrial (CSIC/UPC), C. Llorens Artigas 4-6, 08028 Barcelona, Catalonia 
  (\email{ros@iri.upc.edu}, \url{http://www.iri.upc.edu/people/ros/}).}}

\usepackage{amsopn}

\makeatletter
\newcommand*{\addFileDependency}[1]{
  \typeout{(#1)}
  \@addtofilelist{#1}
  \IfFileExists{#1}{}{\typeout{No file #1.}}
}
\makeatother

\ifpdf
\hypersetup{
	pdftitle={How to design, and tune, a computed torque controller: an introduction, with a Matlab example},
	pdfauthor={L. Ros}
}
\fi

\begin{document}
	
	\maketitle
	
	\begin{abstract}
		This note briefly introduces the computed torque control method for trajectory
tracking. The method is applicable to fully actuated robots, i.e, those whose
inverse dynamics can be solved for any feasible acceleration. This includes
many systems, like robot arms or hands, or any tree-like mechanism with all
its joints actuated. Using simple explanations, we see how such a controller
can be obtained using feedback linearization, and how its gains can be tuned
to satisfy a desired settling time for the error signal. We end up discussing
the advantages and shortcomings of the controller. A companion Matlab script
can be downloaded from \href{https://bit.ly/3QShxYi}{https://bit.ly/3QShxYi}
that implements and tests the controller on a simple actuated pendulum.
	\end{abstract}
	
	
	\section{Goal}
	
\begin{figure}[h!]
		\centering
		\includegraphics[width=.9\linewidth]{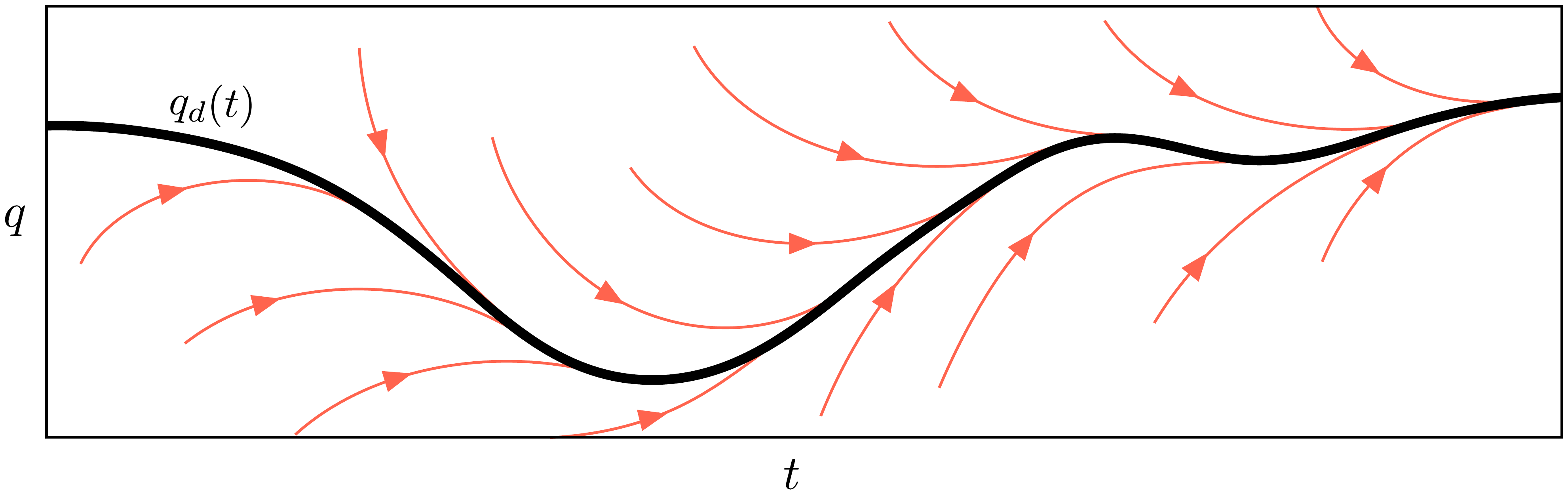}
		\caption{\small Idea of trajectory stabilization: Given a desired trajectory $\vr{q}_d(t)$ (in black), we wish to design a control law that makes the possible real trajectories (in red) asymptotically convergent to $\vr{q}_d(t)$ as $t$ increases. Each red trajectory shown here corresponds to a different set of initial conditions $(t,\vr{q})$ of the robot. Only the behaviour of one component of $\vr{q}$ is depicted.}
		\label{fig:stabilize}
	\end{figure}

	We wish to design a control law that is able to stabilize a robot along a desired
trajectory. We assume the robot is fully actuated, and unconstrained, so its
configuration is given by a vector $\vr{q}=(q_1,\ldots,q_n)$ of $n$ independent
coordinates, each corresponding to an actuated joint angle or
displacement\footnote{Computed-torque controllers can also be designed for
	constrained systems \cite{bordalba2021thesis}, as long as they are fully
	actuated, but these are left out of the scope of this note.}. We also assume
that the trajectory is expressed as a function $\vr{q}_d(t)$
that gives, for each time $t$, the desired value of $\vr{q}$
(Fig.~\ref{fig:stabilize}). The first and second derivatives of this function,
$\vr{\dot{q}}_d(t)$ and $\vr{\ddot{q}}_d(t)$, provide the desired velocity and
acceleration at $t$.
	
	\section{The control law}

	To design the controller, we depart from the Euler-Lagrange equation of the
	robot,
	\begin{equation}
		\mt{M} \; \ddq + \mt{C} \; \dq + \G - \vr{u}_f = \vr{u},
		\label{eq:dynamics}
	\end{equation}
	where $\mt{M}$ is the $n \times n$ positive-definite mass matrix, $\mt{C}$ is
	the $n \times n$ Coriolis matrix, $\G \in \R{n}$ is the gravity term, $\vr{u}_f
	\in \R{n}$ is the generalized force of friction, and $\vr{u} \in \R{n}$ is the
	vector of motor forces and torques ($u_i$ is the force or torque applied by the
	motor at $q_i$).
	
	Our first step will be to change the dynamics of the system into a simpler one
	that is easier to control. This can be done using the feedback law
	\begin{equation}
		\vr{u}= \mt{M} \; \vr{v} + \mt{C} \; \dq + \G - \vr{u}_f,
		\label{eq:feedback1}
	\end{equation}
	where $\vr{v} \in \mathbb{R}^n$ is a new control input that is yet to be chosen.
	Note that if we substitute Eq.~\eqref{eq:feedback1} into~\eqref{eq:dynamics} we
	obtain
	\begin{equation*}
		\mt{M} \; \ddot{\vr{q}} = \mt{M} \; \vr{v}
		\label{eq:eq_u2withMbar}
	\end{equation*}
	and since $\mt{M}$ is nonsingular, this implies that
	\begin{equation}
		\ddot{\vr{q}} = \vr{v}.
		\label{eq:linear}
	\end{equation}
Therefore, the use of the feedback in \cref{eq:feedback1} converts our robot
into a linear system---the double integrator in \eqref{eq:linear}---so we can
now stabilize this system along $\q_d(t)$ using linear control theory. The
process of transforming \eqref{eq:dynamics} into \eqref{eq:linear} is known as
feedback linearization.
	
Note that \eqref{eq:linear} contains $n$ scalar ODEs of the form $\ddot{q}_i =
v_i$, so, if we choose $v_i$ to depend only on $q_i$ and $\dot{q}_i$, the
evolution of $q_i$ will be fully determined by the $i$th ODE $\ddot{q}_i =
v_i$. To stabilize \eqref{eq:linear} along $\q_d(t)$, thus, it
suffices to design a control law for each scalar subsystem $\ddot{q}_i = v_i$
independently. For ease of notation, let us drop the subindex $i$ and write
$\ddot{q}_i = v_i$ simply as
	\begin{equation}
		\ddot{q} = v.
		\label{eq:scalar}
	\end{equation}

	A simple law that stabilizes the system along $q_d(t)$ is
	\begin{equation}
		v = \ddot{q}_d - k_p (q - q_d) - k_v (\dot{q}-\dot{q}_d)
		\label{eq:scalar_law}
	\end{equation}
	where $k_p > 0$ and $k_v > 0$. Certainly, if we substitute \eqref{eq:scalar_law} into \eqref{eq:scalar} we obtain
	\begin{equation}
		\ddot{q} - \ddot{q}_d + k_p (q - q_d) + k_v (\dot{q}-\dot{q}_d) = 0,
		\label{eq:scalar_ode_q}
	\end{equation}
	or, equivalently,
	\begin{equation}
		\ddot{\varepsilon} + k_v \dot{\varepsilon} + k_p \varepsilon = 0,
		\label{eq:scalar_ode_error}
	\end{equation}
	where 
	\begin{equation*}
	\varepsilon(t) = q(t) - q_d(t)	
	\end{equation*}
	is the trajectory error. Equation \eqref{eq:scalar_ode_error} is a 2nd order linear homogeneous ODE that determines the evolution of $\varepsilon(t)$, and it is well known that, if we choose $k_p > 0$ and $k_v > 0$, then
	\begin{equation*}
		\lim_{t\rightarrow\infty} \vr{\varepsilon}(t) = 0
		\label{eq:error_limit}
	\end{equation*}
	independently of the initial conditions $q(0)$ and $\dot{q}(0)$. Such a controller, thus, achieves global stability.
	
	A common choice is to set
	\begin{align*}
		k_p & = \omega_0^2, \\
		k_v & = 2\omega_0,
	\end{align*}
	so \eqref{eq:scalar_ode_error} describes a critically damped harmonic oscillator for the $\varepsilon$ coordinate, with natural frequency $\omega_0$ \cite{feynman2022harmonic,thornton2021classical}. A proper value for  $\omega_0$ can be chosen by specifying the desired settling time $T_s$ for $\varepsilon(t)$. This is the time needed to ensure $\varepsilon(t) \leq 0.02 \cdot \varepsilon(0)$ for $t>T_s$, assuming $\dot{\varepsilon}(0)=0$. Appendix~\ref{ap:set} shows that, to achieve such a time, one must set
	\begin{equation}
		\omega_0 = 5.8339/T_s. 
	\end{equation}
	
\begin{figure}[t!]
		\centering
		\includegraphics[width=\linewidth]{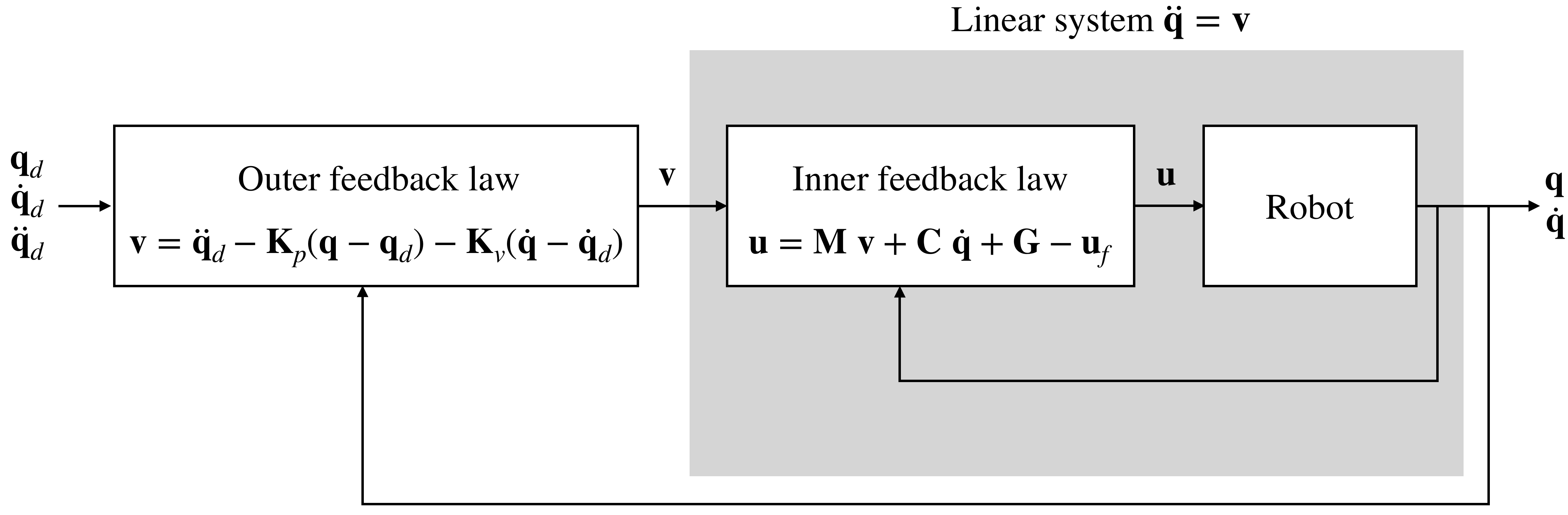}
		\caption{\small Block diagram of the computed torque control law.}
		\label{fig:ctc_diagram}
	\end{figure}

	By particularizing Eq.~\eqref{eq:scalar_law} for each one of the coordinates in \mbox{$\vr{q} = (q_1,\ldots,q_n)$}, we see that the $n$-dimensional control law that stabilizes \eqref{eq:linear} along $\vr{q}_d(t)$ is
	\begin{align}
		\begin{split}
			\vr{v} = \underbrace{\left[
				\begin{array}{c}
					\ddot{q}_{1,d} \\
					\vdots         \\
					\ddot{q}_{n,d}
				\end{array}
				\right]}_{\ddot{\vr{q}}_d}
			- \underbrace{\left[
				\begin{array}{ccc}
					k_{p,1} &        &         \\
					& \ddots &         \\
					&        & k_{p,n}
				\end{array}
				\right]}_{\mt{K}_p} \underbrace{\left[
				\begin{array}{c}
					q_1 - q_{1,d} \\
					\vdots        \\
					q_n - q_{n,d}
				\end{array}
				\right]}_{\vr{q} - \vr{q}_d}
			- \\
			-
			\underbrace{\left[
				\begin{array}{ccc}
					k_{v,1} &        &         \\
					& \ddots &         \\
					&        & k_{v,n}
				\end{array}
				\right]}_{\mt{K}_v} \underbrace{\left[
				\begin{array}{c}
					\dot{q}_1 - \dot{q}_{1,d} \\
					\vdots                    \\
					\dot{q}_n - \dot{q}_{n,d}
				\end{array}
				\right]}_{\dot{\vr{q}} - \dot{\vr{q}}_d}
			\label{eq:feedback2}
		\end{split}
	\end{align}
	where $k_{p,i}$ and $k_{v,i}$ are the position and velocity gains for $q_i$, and $q_{i,d}$ is the $i$th coordinate of $\q_d$.
	If we write Eq.~\eqref{eq:feedback2} as
	\begin{equation*}
		\vr{v} = \ddq_d - \mt{K}_p \left(\q - \q_d \right) - \mt{K}_v (\dq - \dq_d)
		\label{eq:accel}
	\end{equation*}
	and substitute it back into Eq.~\eqref{eq:feedback1}, we finally obtain
	\begin{equation}
		\vr{u} = \mt{M} \;
		\underbrace{
			\big[ \;
			\ddq_d - \mt{K}_p ( \q - \q_d )
			-
			\mt{K}_v (\dq - \dq_d ) \;
			\big]
		}_{\vr{v}}
		+
		\mt{C} \; \dq + \G - \vr{u}_f
		\label{eq:ctc_law}
	\end{equation}
	which is the usual computed-torque control law for a fully actuated robot (Fig.~\ref{fig:ctc_diagram}).
	
	Observe that, to control the robot using \eqref{eq:ctc_law}, we need sensory feedback of the $\vr{q}$ and $\dq$ values at each iteration of the control loop, not only to obtain the configuration and velocity errors $\q - \q_d$ and $\dq - \dq_d$, but also the values of $\mt{M}$, $\mt{C}$, $\G$, and $\vr{u}_f$, whose dependencies on $\q$ and $\dq$ are
	\begin{equation*}
		\mt{M}(\q), \ \ \mt{C}(\q,\dq), \ \ \G(\q), \ \ \vr{u}_f(\q,\dq).
	\end{equation*}
	Moreover, the calculation of $\vr{u}$ using \eqref{eq:ctc_law} has to be fast to achieve a high-enough control frequency. Since this calculation is equivalent to solving the inverse dynamics for the $\vr{v}$ value in \eqref{eq:ctc_law}, achieving fast algorithms for this purpose is of utmost importance in this context.
	
	\section{Discussion}
	
	The computed torque control law in Eq. \eqref{eq:ctc_law} ensures the
convergence of the robot to $\q_d(t)$ independently from the initial conditions
$\vr{q}(0)$. This property is remarkable, and it is sometimes referred to as
``global stability''. The main drawback of the law is that it requires a very
good model of the robot, otherwise the feedback in Eq. \eqref{eq:feedback1}
will not transform the system into a double integrator as required. Another
drawback is that limits in the control forces or torques are not considered in
this method. If such limits are somehow surpassed while operating the robot,
the convergence towards $\vr{q}_d(t)$ may take longer, or even be impossible.
This is why robots being controlled by such laws typically employ powerful
motors. Also, choosing a suitable settling time $T_s$ is crucial. Small $T_s$
values are always preferrable, but yield a more aggressive
controller, which increases the risk of violating the force limits. Thus,
$T_s$ cannot be chosen too small to ensure a proper functioning of the robot.
	
\section{An example} A companion Matlab script implements and tests the
controller on a simple actuated pendulum~\cite{ros2023ctcscript}. In its default
run, the script starts showing the pendulum at rest in its bottom
configuration, and then simulates the tracking of a step trajectory $\vr{q}_d(t)$
that requires the pendulum to go to its upright configuration, and then back to
the initial one. A perturbation force is applied while the pendulum is in
inverted balance to see how the controller is able to counteract disturbances.
The script is fully commented for ease of comprehension.
	
	\section{Further reading} Many books cover the topic of this note. For further
	details we refer to
	\cite{spong2020robot,lynch2017modern,slotine1991applied,khatib2016springer}. The book in \cite{spong2020robot} covers the advanced topic of designing a computed torque controller that is more robust to various sources of uncertainty, such as modeling errors, unknown loads, or computational errors.	Note also that, like \cite{spong2020robot}, some references use the term ``inverse dynamics control'' as a synonym of ``computed torque control''.
	
	\section*{Acknowledgements}
	I am grateful to Enric Celaya, Ricard Bordalba, Pere Gir\'o, Adriano del R\'{\i}o, and Siro Moreno for fruitful discussions around the topic of this note.


	\bibliographystyle{siamplain}
	\bibliography{references}

	\newpage
	\appendix
	
	\section{Natural frequency for a desired settling time}
	\label{ap:set}
	\begin{figure}[b!]
		\centering
		\includegraphics[width=0.5\linewidth]{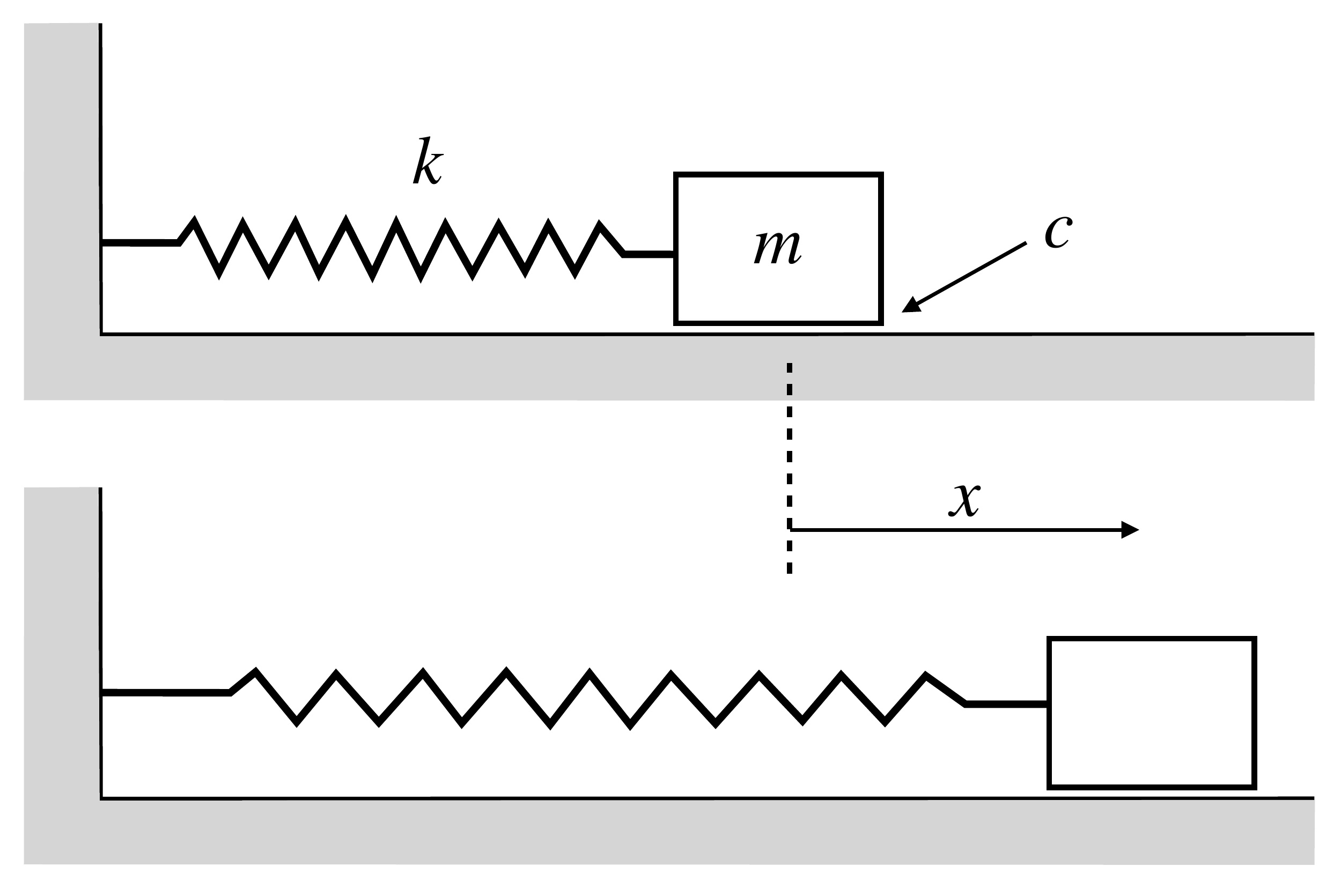}
		\caption{\small A harmonic oscillator with viscous friction at the ground contact. Top: the equilibrium configuration. Bottom: a generic configuration at a distance $x$ from the equilibrium one.}
		\label{fig:mass_spring}
	\end{figure}
	Consider a damped harmonic oscillator described by the ODE
	\begin{equation}
		m \ddot{x} = -k x - c \dot{x},
		\label{eq:ho}
	\end{equation}
	where $m$ is the attached mass, $x$ is the mass position, $k$ is the spring constant, and $c$ is the viscous friction coefficient of the mass-ground contact (Fig.~\ref{fig:mass_spring}). By defining
	\begin{gather}
		\omega_0 = \sqrt{\frac{k}{m}} \\
		\xi = \frac{c}{2\sqrt{mk}}
	\end{gather}
	Eq. \eqref{eq:ho} can be written as
	\begin{equation}
		\ddot{x} + 2 \xi \omega_0 \dot{x} + \omega_0^2 = 0,
		\label{eq:2nd_order}
	\end{equation}
	where $\omega_0$ is the natural frequency of the oscillator and $\xi$ is its damping ratio. It is well known that, for a fixed value of $k$ and $m$, the fastest way in which $x$ converges to zero, without oscillations, occurs when $\xi = 1$. In such a case,  \cite{thornton2021classical,tenenbaum1985ordinary} show that the solution of \eqref{eq:2nd_order} is
	\begin{equation}
		x(t) = (c_1+c_2 t) e^{-\omega_0 t},
		\label{eq:sol_critdamp}
	\end{equation}
	where $c_1$ and $c_2$ are to be determined using the initial conditions
	\begin{gather}
		x(0) = x_0, \label{eq:ic1} \\
		\dot{x}(0) = v_0. \label{eq:ic2}
	\end{gather}
	Evaluating \eqref{eq:ic1} and \eqref{eq:ic2} using \eqref{eq:sol_critdamp}, and solving for $c_1$ and $c_2$, we obtain
	\begin{gather}
		c_1 = x_0, \\
		c_2 = v_0 + x_0 \omega_0,
	\end{gather}
	so \eqref{eq:sol_critdamp} can be written as
	\begin{equation}
		x(t) = \left[\:x_0 + (v_0 + x_0 \omega_0) \; t \: \right] \cdot e^{-\omega_0 t}.
	\end{equation}
	
	We wish to find the value $\omega_0$ for which $x(t)$ achieves a desired settling time $T_s$. This is the time needed for $x(t)$ to be below $0.02 \cdot x_0$, assuming $v_0 = 0$ and $x_0 \neq 0$. Thus, we must solve
	\begin{equation}
		(x_0 + x_0 \omega_0 T_s) e^{-\omega_0 T_s} = 0.02 \: x_0,
		\label{eq:ineq}
	\end{equation}
	or equivalently, dividing both sides by $x_0$,
	\begin{equation}
		(1 + \omega_0 T_s) e^{-\omega_0 T_s} = 0.02.  \label{eq:ineq2}
	\end{equation}
	If we define $P = \omega_0 T_s$, and rearrange the terms, \eqref{eq:ineq2} becomes
	\begin{equation}
		(1 + P) = 0.02 \: e^{P}
		\label{eq:ineq3}
	\end{equation}
	which has the single solution
	\begin{equation}
		P \approx 5.8339
		\label{eq:Pvalue}
	\end{equation}
	for $P >0$ ($P$ must be positive because $\omega_0 > 0$ and $T_s > 0$). Therefore, the value $\omega_0$ that achieves the desired settling time is
	\begin{equation}
		\omega_0 \approx \frac{5.8339}{T_s}.
		\label{eq:omega0Ts}
	\end{equation}
	
	%

\end{document}